\documentclass{article}

\usepackage{verbatim}
\usepackage[hidelinks]{hyperref}
\usepackage{url}
\usepackage{amsmath}
\usepackage{rotating}
\usepackage{bbm}
\usepackage{multirow}
\usepackage{caption}
\usepackage{subcaption}
\usepackage{graphicx}
\usepackage{hyperref}
\usepackage[table]{xcolor}
\usepackage{wrapfig}  
\usepackage[font=footnotesize]{caption}
\usepackage[final]{corl_2024} 

\title{Versatile and Generalizable Manipulation via Goal-Conditioned Reinforcement Learning with Grounded Object Detection}

\author{Huiyi Wang \& Colin Bellinger \\
Digital Technologies,
National Research Council of Canada,
Ottawa, Canada \\
\texttt{\{huiyi.wang, colin.bellinger\}@nrc-cnrc.gc.ca} \\
\And
Fahim Shahriar \&  Alireza Azimi \& Gautham Vasan \& Rupam Mahmood\\
University of Alberta, Amii \\
Edmonton, Canada\\
\texttt{\{fshahri1, sazimi, vasan, ashique\}@ualberta.ca}
}

\begin{document}
\maketitle

\vspace{-1.cm}
\begin{abstract}
General-purpose robotic manipulation, including reach and grasp, is essential for deployment into households and workspaces involving diverse and evolving tasks. Recent advances propose using large pre-trained models, such as Large Language Models and object detectors, to boost robotic perception in reinforcement learning. These models, trained on large datasets via self-supervised learning, can process text prompts and identify diverse objects in scenes, an invaluable skill in RL where learning object interaction is resource-intensive. This study demonstrates how to integrate such models into Goal-Conditioned Reinforcement Learning to enable general and versatile robotic reach and grasp capabilities. We use a pre-trained object detection model to enable the agent to identify the object from a text prompt and generate a mask for goal conditioning. Mask-based goal conditioning provides object-agnostic cues, improving feature sharing and generalization. The effectiveness of the proposed framework is demonstrated in a simulated reach-and-grasp task, where the mask-based goal conditioning consistently maintains a $\sim$90\% success rate in grasping both in and out-of-distribution objects, while also ensuring faster convergence to higher returns.
\end{abstract}



\section{Introduction}

\vspace{-0.09cm}

Learning to reach and grasp objects with general robotic systems using reinforcement learning (RL) requires the agent to learn to recognize each object to determine the appropriate reaching strategy. This process is not only time-consuming but also potentially expensive in terms of data acquisition. Moreover, such systems often struggle to generalize to objects that were not included in the training distribution. Recently, the large pre-trained models have shown the potential to improve sample efficiency and enhance contextual understanding in reinforcement learning through the use of shaped reward by language-driven prompts \citep{kwon2023reward, palo2023towards}. GroundingDINO \citep[G.DINO, ][]{GroundingDINO} is a Grounded Segment Anything (SAM) model trained via self-supervisions on large public data to achieve accurate object detection and segmentation from text prompts. Hence, we postulate that coupling RL agent decision-making with pre-trained object detectors and an appropriate abstraction could reduce the burden of target object recognition in the agent, a task that is more efficiently learned offline from a static independent and identically distributed (IID) dataset. This capability is crucial for handling open-world scenarios, enabling reinforcement learning agents to effectively interact with previously unseen objects. 

In this work, we utilize Goal-Conditioned RL (GCRL) to learn a single policy for reaching and grasping, where the target object is specified via a text prompt, such as `apple on the right', at the start of each episode. We propose specifying the goal condition to the GCRL agent as an abstract target object mask, that can be generated by a grounded object detector. We show that this method, in combination with tactile-visual sensing in the form of reward, allows us to achieve robotic reach-and-grasp tasks in a versatile framework that generalization for out-of-distribution objects. The mask-based goal conditioning offers several advantages. First, It provides a \textit{relative} goal location with respect to the agent's current observational, dynamically adjusting throughout its interactions with the environment. Second, it enables efficient feature sharing and flexibility in adapting the trained policy to novel goal objects or locations. This is because mask-based goal conditioning abstracts the specific details of the target object, enabling the reaching component to be learned more efficiently and with greater transferability. This approach allows the learning to reach to be partially independent of any particular object. Third, It eliminates the need for a significant amount of experience in video datasets of human demonstrations like that required in Large Language Model (LLM)  \citep[e.g., Reusable Representation for Robotic Manipulations (R3M), ][]{nair2022rm}. Finally, it has a lower dimension than the raw RGB image of the goal and is both financially and computationally less expensive for inference than LLMs, facilitating faster training.

\begin{wrapfigure}{r}{0.43\textwidth}
    \centering
    \vspace{-0.1cm}
    \begin{subfigure}{.44\textwidth}
        \includegraphics[scale=0.33]{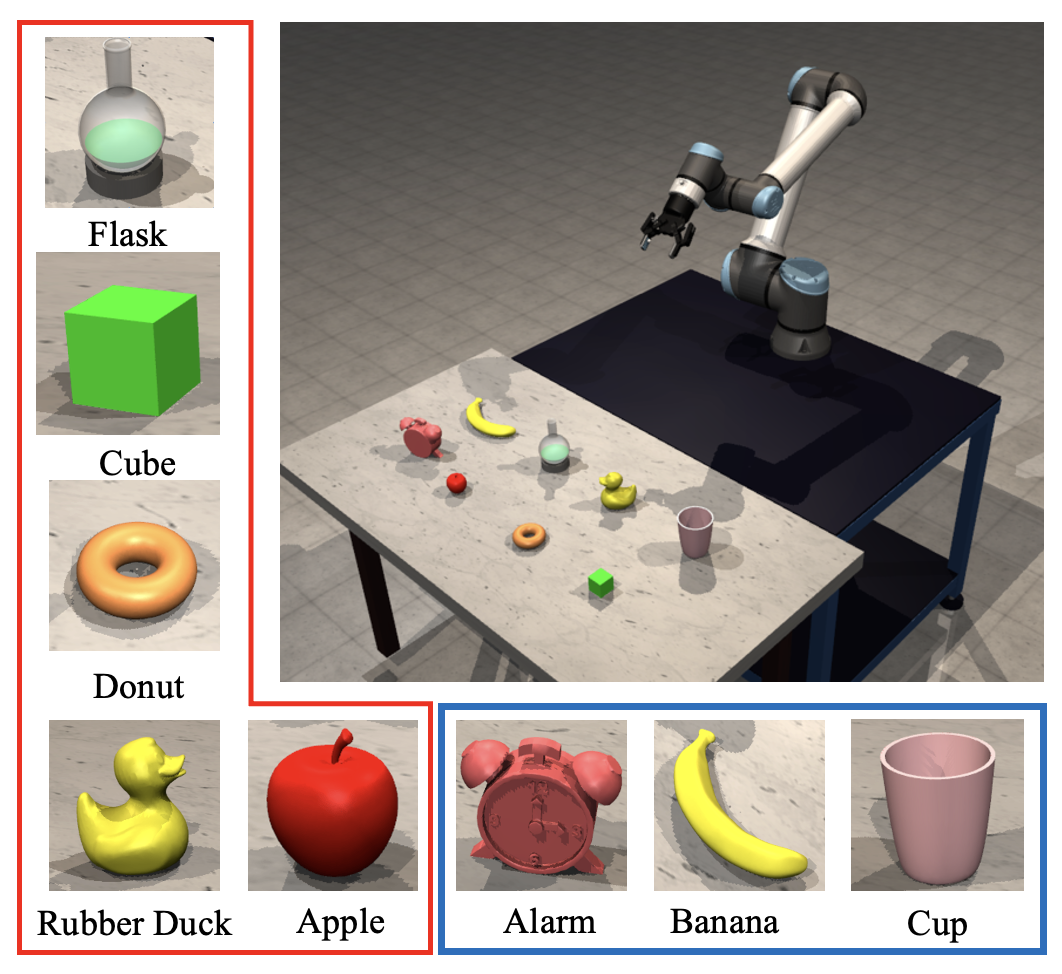} 
    \end{subfigure}
    \footnotesize
    \caption{Simulation setup of reach and grasp, with a UR10e robot plus 2F-85 robotiq gripper with 7 distinct objects chosen from object$\_$sim \citep{dasari2023pgdm} and flask created with Blender \citep{Blender}. The in-distribution training is bounded in red, while the out-of-distribution testing objects are bounded in blue.}
    \label{fig: setup}
    \vspace{-0.5cm}
\end{wrapfigure}

\section{Proposed Method Formalism}
\label{MC-GCRL}
\vspace{-0.09cm}

We proposed the use of a pre-trained grounded object detector during perception that allows the agent to utilize a text input of the target provided by the environment to generate a target object mask for goal conditioning. This mask is updated at each time step based on the agent's ego-centric observation. To demonstrate the effectiveness of the target masks for goal conditioning in a reach-and-grasp task, we test the proposed method based on oracle-generated masks and masks generated from the output of pre-trained grounded object detectors, G.DINO \footnote{We used G.DINO to generate the BB, though other pre-trained models could also serve this function.}.

At the beginning of each episode, the environment provides a text string specifying reaching and grasping goal for the episode. At each timestep, the target object text description and a copy of the current ego-centric observation are mapped to a bounding box (BB) generated around the target object in the current frame. A one-channel image is created from this where all objects outside the BB are black and those inside the BB are white. The masking process is defined as
$g_m(t) = E(o_i(t))$.
where $E$ encompasses the process of (1) using a model to identify the goal object by text strings, and (2) generating a mask corresponding to the area of the bounding box. Here, $o_i$ is the image observation at each timestep $t$. In this work, $o_i$ is an egocentric view from a camera on the end effector of a robotic arm. The agent selects the next action based on the current image and proprioception observation, $o_t$, and goal condition mask, $g_m$: $\pi(a_{t+1} | o_t, g_m)$.

\section{EXPERIMENTAL SETUP}
\vspace{-0.09cm}

We cross-compared the proposed mask-based goal conditioning with standard goal conditioning using vector or image in terms of generalization over five in-distribution objects and evaluated on three out-of-distribution objects trained using PPO (see Figure \ref{fig: setup}). Here, we use a distance-based reward in order to focus solely on the goal-conditioning. The goal conditioning setups are (see image demonstration in Figure \ref{fig:goal}, Appendix):

\noindent 1) \textit{Vector-based goal-conditioning:} a one-hot encoding of the 8-element array. This provides space for the five training objects and three out-of-distribution objects. 

\noindent 2) \textit{Image-based goal-conditioning:} A $3 \times 224 \times 224$ pixel generic image of the goal object that is selected at the start of each episode is appended to the observation at each timestep (Figure \ref{fig: setup}). 

\noindent 3) \textit{Mask-based goal-conditioning:} A $1 \times 224 \times 224$ binary pixel mask of the target object. The experiments include two setups: \textit{i}) ground truth (GT) target object masks generated by a bounding box (BB) oracle, and \textit{ii}) masks generated from BB inferences produced by G.DINO using the text specification of the goal object. The ground-truth BBs are generated within the MuJoCo simulation by transforming the object's coordinates into pixel points on the camera's view.

\section{EXPERIMENTAL RESULTS}
\vspace{-0.09cm}
\textbf{Generalizability of Object Masks for GCRL}

\begin{figure}[h]
    \centering
    \includegraphics[width= 0.99\linewidth]{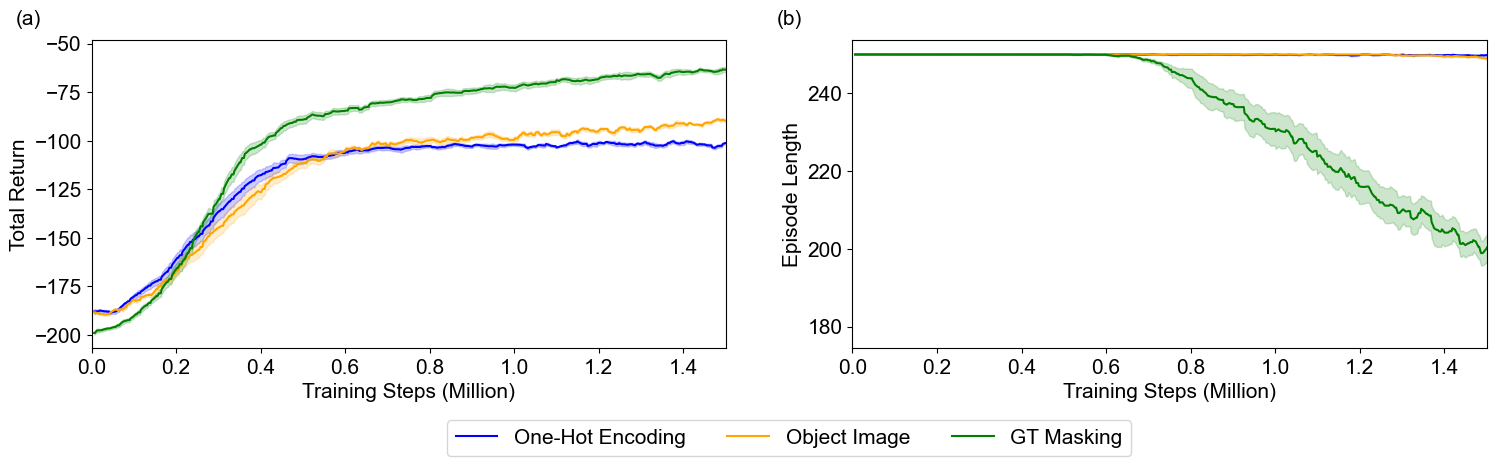}
    \caption{Comparison of learn with ground truth (GT) mask for goal conditioning to vector and image-based methods for a reach-and-grasp task reported with standard error.}
    \label{fig:reach_plot}
    \vspace{-0.2cm}
\end{figure}

\textit{(1) Learn efficiency with object Masks for GCRL:} We compare the performance between using standard methods of goal conditioning based on one-hot encoding and goal objects against our proposed masking method. The standard methods of goal conditioning achieve sub-optimal returns, as shown in blue and orange in Figure \ref{fig:reach_plot}(a).  In comparison, our methods, as shown in the green of Figure \ref{fig:reach_plot}(a), converge faster to a $\sim 25\%$ higher return and learn to successfully grasp the object. Additionally, Figure \ref{fig:reach_plot}(b) shows that mask-based goal conditioning learns a policy that successfully grasps in significantly fewer steps than the maximum episode length, whereas the others do not.

\def\arraystretch{1.1}
\begin{wraptable}{r}{7.cm}
\footnotesize
\caption{Comparison of goal conditioning methods and their grasping success rates (* here during evaluation, single-gripper contact is used as the success condition) for in-distribution and out-of-distribution objects.  }
\label{table: env}
\centering
\begin{tabular}{|p{3.cm}|c|c|}

\hline
 \multirow{2}{*}{\centering \textbf{Goal Conditioning}}  & \multicolumn{2}{c|}{Grasping Success Rate}\\
\cline{2-3}
 & In-distri & Out-distri\\
\hline
 One-Hot Encoding   & 0.13 & 0.2 \\
\hline
 Object Image  & 0.62 & 0.28\\
\hline
 GT Masking  & 0.89 & 0.9 \\
\hline
\end{tabular}
\vspace{-0.2cm}
\end{wraptable}

\textit{(2) Generalization with Object Masks for GCRL:} 
Our mask-based goal conditioning also shows robustness in both in and out-of-distribution (OOD) objects. 
When using a general goal image for goal conditioning, although the in-distribution object grasping success rate reaches $62\%$, it quickly drops to $28\%$ for out-of-distribution objects. This is likely due to the model's inability to learn higher-level feature abstractions that can be shared across different objects in the training set. However, performance might improve with extended training time and the inclusion of more in-distribution objects. Alternatively, our approach demonstrates robustness in grasping, achieving a success rate of 89\% for in-distribution objects and remaining on par at approximately 90\% for OOD objects.

\textbf{Target Masking with G.DINO for GCRL}

Additionally, we demonstrate the use of G.DINO for mask generation in the proposed robotic grasping. This enables the agents to utilize knowledge from the pre-trained model in the observation image instead of information on the location of the target object. The results are summarized in Table \ref{table: GDINO_EVA}. We analyze the grasping success rates when using G.DINO to generate masks during evaluation, comparing policies trained with G.DINO-inferenced masks versus those using GT masks. Our results indicate a higher grasping success rate with policies trained with GT masks compared to those trained on G.DINO-generated masks for in-distribution objects. This discrepancy is primarily attributed to the inherent noise in G.DINO inferences that can incorrectly identify the target object, leading to masks that direct the agent toward an erroneous object.

To further explore the impact of noise on G.DINO inferences, we analyze the agent's performance with out-of-distribution objects in three scenarios: (1) the goal object alone, randomly positioned; (2) with one distractor object; and (3) with two distractor objects.  We observe that when only the goal object is presented, the grasping rate is the highest (Table \ref{table: GDINO_EVA}), illustrating that the presence of additional objects introduces noise and complicates G.DINO's ability to accurately identify the target object. Notably, when only the target object is present using a policy trained with GT masking, the agent achieves a success rate of approximately 82\%.

\def\arraystretch{1.2}
\begin{table}
\footnotesize
\caption{Grasping success rate in an environment with G.DINO inferred mask for goal conditioning evaluated with policies trained with either G.DINO (GD) or Ground Truth (GT) Masking. For out-of-distribution objects, we evaluate the grasping success rate when 1, 2, or 3 objects are presented on the table.  }
\label{table: GDINO_EVA}
\centering
\begin{tabular}{|l|c|c|c|c|}

\hline
\multirow{3}{*}{\centering \textbf{Policy}} & \multicolumn{4}{c|}{Grasping Success Rate}\\
\cline{2 - 5}
& \multirow{2}{*}{In-distribution} & \multicolumn{3}{c|}{Out-of-distribution} \\
\cline{3 - 5}
 & & 1 &2 &3 \\
\hline
\hline
Train with GD, evaluate with GD & 0.21 & 0.28 &0.22 & 0.24\\
\cline{1 - 5}
Train  with GT, evaluate with GD  & 0.9 & 0.82 & 0.79 & 0.67 \\
\hline
\end{tabular}
\vspace{-.2cm}
\end{table}

\section{DISCUSSION}
\vspace{-0.09cm}
The choice of using G.DINO was made by comparing it with other open-vocabulary object detection. In comparison to GLIP, a phrase grounding method that involves associating phrases with corresponding visual cues \citep[][]{Li2022}, G.DINO outperforms in open-set object detection \citep[][]{Firoozi2023}. Additionally, G.DINO allows zero-shot inferences and has higher mean averaged precision than YOLOv8 \citep[][]{Son2024}. The grasping success rate of using G.DINO inference is only $\leq 60\%$ of the runs using the GT masking (Table \ref{table: GDINO_EVA}). The more objects that are presented on the table, the lower the grasping success rate, as the pre-trained model has a higher chance of generating false positive target masking. Additionally, the accuracy of the BB model decreases accordingly when the object's completeness, angle, and distance are changed w.r.t. an ego-centric camera. 

Another limitation of G.DINO is its significant time cost during inference loops in comparison to other pre-trained models such as YOLO \citep[][]{Son2024}. To alleviate this issue, we consider the use of asynchronous learning \citep{Gu2016, Yuan2022} for real-time inference as part of our future work. This method has been demonstrated by \citet{Yuan2022} to substantially outperform sequential learning, particularly when learning updates are computationally expensive.

\section{CONCLUSION}
\vspace{-0.09cm}
In this work, we proposed the use of a grounded object detection in robotic perception in combination with tactile-visual sensing reward in goal-conditioning learning to achieve generalizable robotic reach-and-grasp. We employ a pre-trained object detection model, GroundingDINO, to generate a bounding box around the goal object, which is transformed into a binary mask that feeds into the observation for goal conditioning. We evaluated our framework on a reach-and-grasp task with a simulated UR10e robotic arm. The results demonstrated that our proposed framework enables more efficient feature sharing across multiple goal objects and allows robust generalization and faster convergence to out-of-distribution objects, outperforming traditional goal conditioning like one-hot encoding or generic object images.


\clearpage
\acknowledgments{}

This research is funded through the National Research Councils of Canada’s Artificial Intelligence for Design Challenge program. 


\bibliography{example}  

\appendix
\section{Appendix}

\subsection{Experimental Setup}
\textbf{UR10e Goal Conditioned Reaching and Grasping Environmnet:} Our simulated environment includes a UR10e robotic arm with a 2F-85 gripper with 7 degrees of freedom in the MuJoCo simulator \citep{Todorov2012}. The environment has a multi-input observation space composed of the $3 \times 224 \times 224$ RGB image from the end-effector camera and UR10e 7D proprioception.

The target objects are placed on a table in front of the UR10e robotic arm and within the initial view of the end-effector camera. At the start of each episode, the positions of the five objects are randomly swapped, and they are collectively translated by a small, random distance along the $x$ and $y$ axes. The task is considered successfully completed when both pads of the gripper make contact with the goal object. The maximum length of the episode is set to 250 steps. 

\textbf{Algorithms and Evaluation:} For our experiments, we train the agent using an on-policy algorithm, Proximal Policy Optimization \cite[PPO, ][]{PPO} implemented in stable-baselines3 \citep{SB3}. The PPO hyper-parameters are included in Appendix \ref{hyper}.

Image resolution has a strong influence on the accuracy of G.DINO. Hence, the inference model receives higher resolution---$3 \times 800 \times 800$ pixels---images, whereas the RL policy is limited to $3 \times 224 \times 224$ pixels. We set the G.DINO inference threshold to 0.55 to balance the true and false positive rates.

We evaluate the proposed mask-based goal-conditioning strategy in terms of the mean and standard error of the return and episode length averaged over 10 random seeds, as well as the reaching and grasping success rate on in- and out-of-distribution objects. For the evaluation of the grasping success rate of the optimally seeded policy, we define successful grasping based on a criterion of single gripper contact.

\subsection{Goal Conditioning Setup}
\begin{figure}[ht]
    \centering
    \includegraphics[width=0.9\linewidth]{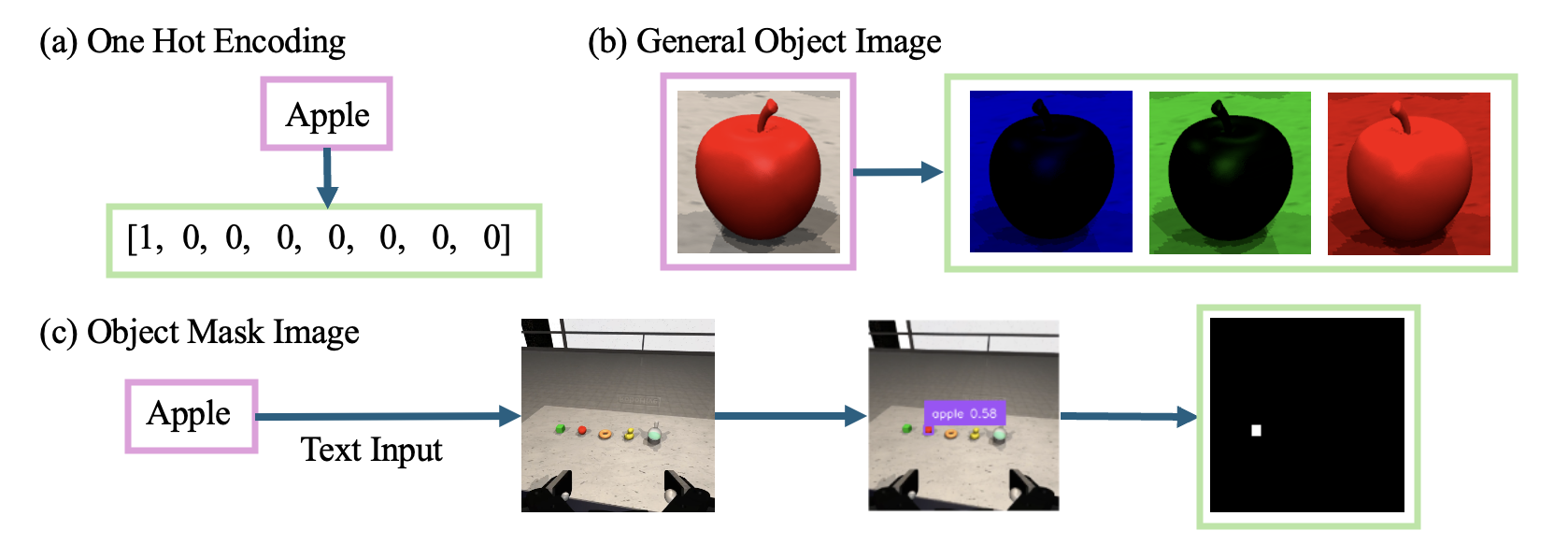}
    \caption{Three different goal conditioning for reach-and-grasp task when apple is chosen as the target object. The green bounding box shows the final goal conditioning representation. }
    \label{fig:goal}
    \vspace{-0.2cm}
\end{figure}

\newpage
\subsection{Grasping Demonstration with Different Goal Conditioning}

\begin{figure}[ht]
    \centering
    \vspace{-0.1cm}
        \includegraphics[scale=0.65]{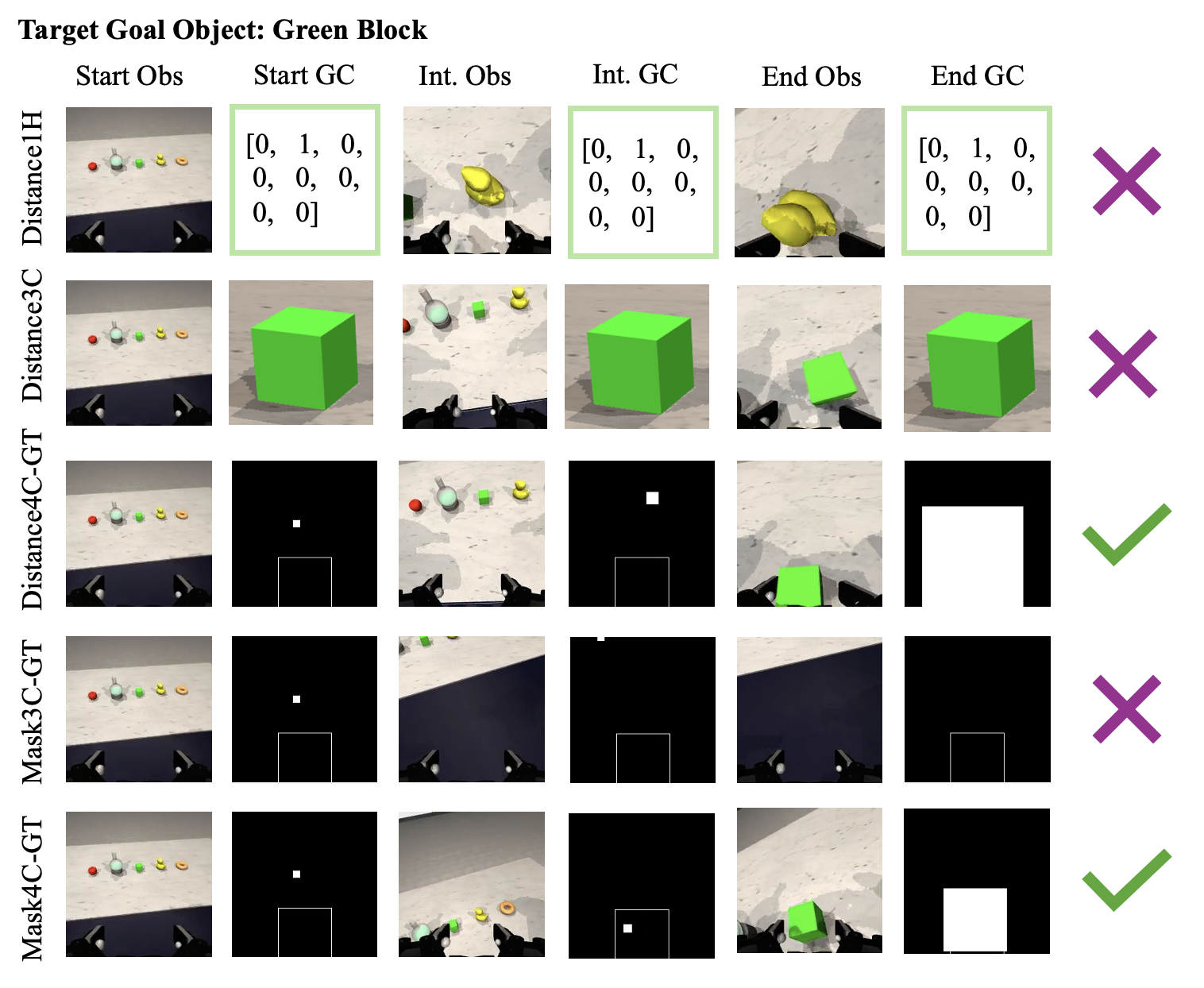} 
    \caption{Visualization of the goal conditioning (GC) and observation during the start, intermediate, and ending steps of the episode. The last column indicates whether or not the reach-and-grasp is successful. }
    \label{fig: results}
    \vspace{-0.0cm}
\end{figure}

\subsection{Hyperparameter Tuning} \label{hyper}
We present here the choice of hyperparameter for different goal conditioning. The learning rate and clip range were determined after a hyperparameter sweep, ranging from [1e-4, 5e-4] and [0.0, 0.1] respectively to avoid slow convergence and catastrophic unlearning, which is a common phenomenon in PPO. We experimented with both constant and linearly scheduled decreasing rates.  The entropy coefficient was set at 0.01 to balance exploration and exploitation.  The neural network size is determined by the channel dimensions of the images processed during the RL training loop. Using either a one-hot encoding or a mask image for goal conditioning requires a smaller neural network to achieve a robust policy. We set the maximum episode length at 250 to allow the agent sufficient time to fully explore the table area while maintaining efficient training sessions.

\def\arraystretch{1.2}
\begin{table}[ht]
\caption{Hyperparameters of the PPO neural Network. GT = Ground Truth Masking, GD = GroundingDINO Masking, ls($x$) = linear schedule of decrease from $x$ to 0 over the entire training steps. }
\label{table: hyper}
\centering
\begin{tabular}{l|| c|c |c|c}
 \textbf{Hyperparameter}& 
\textit{Distance-1H}& \textit{Distance-3C} & \textit{Distance-4C-GT} & \textit{Distace-4C-GD} \\
\hline
\hline
Learning Rate ($\alpha$)& ls(3e-4)& ls(2e-4) & ls(3e-4) & ls(2e-4)\\
\hline
Clip Range &  \multicolumn{4}{c}{ls(0.1)}\\
\hline
Entropy Coefficient&  \multicolumn{4}{c}{0.01}\\
\hline
Activation &  \multicolumn{4}{c}{ReLU} \\
\hline
Neural Network Size& 512 & 1024 & 512 & 512 \\
\hline
Max Episode Length & \multicolumn{4}{c}{250}\\
\hline
\end{tabular}
\end{table}

\end{document}